\documentclass[letterpaper, 10 pt, conference]{class/ieeeconf}  

\IEEEoverridecommandlockouts                              
\usepackage{xcolor}
\usepackage{siunitx}  
\usepackage{graphicx}
 \graphicspath{{./}{figures/robot_demonstration}}
\graphicspath{{figures/robot_demonstration/}}

\usepackage{amsmath}
\usepackage{amssymb}
\usepackage{latexsym}
\usepackage{url}
\usepackage{cite}
\usepackage{relsize}
\usepackage{multirow}
\usepackage{afterpage}
\usepackage{ifthen}
\usepackage{graphicx}
\usepackage{algpseudocode,algorithm,algorithmicx}
\usepackage{subfigure}
\usepackage{flushend}
\usepackage{epstopdf}
\usepackage{stackengine}
\usepackage{textcomp} 

\usepackage{gensymb}
\usepackage[bookmarks=false, linkcolor=blue, urlcolor=blue, citecolor=blue]{hyperref} 
\usepackage{booktabs}
\usepackage{makecell}
\usepackage{hhline}
\usepackage{courier}
\usepackage{lipsum}

\usepackage{xcolor} 

\hypersetup{
    colorlinks=true,
    linkcolor=red,
    filecolor=magenta,      
    urlcolor=blue,
    pdfstartview={FitH}    }

\usepackage{xspace}

\providecommand{\kuka}{\textsc{KUKA} LBR iiwa R820\xspace}

\newcommand{\etal}{\textit{et al.}}

\title{\LARGE \bf Open-Vocabulary Affordance Detection in 3D Point Clouds}

\author{Toan Nguyen$^{1,2}$, Minh Nhat Vu$^{3,4}$, An Vuong$^1$, Dzung Nguyen$^1$, Thieu Vo$^5$, Ngan Le$^6$,  Anh Nguyen$^7$
\thanks{$^1$ FPT Software AI Center, Vietnam {\tt \{toannt28, anvd2, dungnt20\}@fpt.com}}
\thanks{$^2$ VNUHCM-University of Science, Ho Chi Minh City, Vietnam}
\thanks{$^3$ Automation \& Control Institute (ACIN), TU Wien, Vienna, Austria {\tt vu@acin.tuwien.ac.at}}
\thanks{$^4$ Center for Vision, Automation \& Control, AIT Austrian Institute of Technology (GmbH), Vienna, Austria {\tt minh.vu@ait.ac.at}}
\thanks{$^5$ Faculty of Mathematics and Statistics, Ton Duc Thang University, Ho Chi Minh City, Vietnam {\tt vongocthieu@tdtu.edu.vn}}
\thanks{$^6$ Department of Computer Science \& Computer Engineering, University of Arkansas {\tt thile@uark.edu}}
\thanks{$^7$ Department of Computer Science, University of Liverpool, UK {\tt anh.nguyen@liverpool.ac.uk}}}

\begin{document}

\newtheorem{problem}{Problem}
\newtheorem{lemma}{Lemma}
\newtheorem{theorem}[lemma]{Theorem}
\newtheorem{claim}{Claim}
\newtheorem{corollary}[lemma]{Corollary}
\newtheorem{definition}[lemma]{Definition}
\newtheorem{proposition}[lemma]{Proposition}
\newtheorem{remark}[lemma]{Remark}
\newenvironment{LabeledProof}[1]{\noindent{\it Proof of #1: }}{\qed}

\def\beq#1\eeq{\begin{equation}#1\end{equation}}
\def\bea#1\eea{\begin{align}#1\end{align}}
\def\beg#1\eeg{\begin{gather}#1\end{gather}}
\def\beqs#1\eeqs{\begin{equation*}#1\end{equation*}}
\def\beas#1\eeas{\begin{align*}#1\end{align*}}
\def\begs#1\eegs{\begin{gather*}#1\end{gather*}}

\newcommand{\poly}{\mathrm{poly}}
\newcommand{\eps}{\epsilon}
\newcommand{\e}{\epsilon}
\newcommand{\polylog}{\mathrm{polylog}}
\newcommand{\rob}[1]{\left( #1 \right)} 
\newcommand{\sqb}[1]{\left[ #1 \right]} 
\newcommand{\cub}[1]{\left\{ #1 \right\} } 
\newcommand{\rb}[1]{\left( #1 \right)} 
\newcommand{\abs}[1]{\left| #1 \right|} 
\newcommand{\zo}{\{0, 1\}}
\newcommand{\zonzo}{\zo^n \to \zo}
\newcommand{\zokzo}{\zo^k \to \zo}
\newcommand{\zot}{\{0,1,2\}}
\newcommand{\en}[1]{\marginpar{\textbf{#1}}}
\newcommand{\efn}[1]{\footnote{\textbf{#1}}}
\newcommand{\vecbm}[1]{\boldmath{#1}} 
\newcommand{\uvec}[1]{\hat{\vec{#1}}}
\newcommand{\thv}{\vecbm{\theta}}
\newcommand{\junk}[1]{}
\newcommand{\var}{\mathop{\mathrm{var}}}
\newcommand{\rank}{\mathop{\mathrm{rank}}}
\newcommand{\diag}{\mathop{\mathrm{diag}}}
\newcommand{\tr}{\mathop{\mathrm{tr}}}
\newcommand{\acos}{\mathop{\mathrm{acos}}}
\newcommand{\atantwo}{\mathop{\mathrm{atan2}}}
\newcommand{\SVD}{\mathop{\mathrm{SVD}}}
\newcommand{\quadf}{\mathop{\mathrm{q}}}
\newcommand{\linterp}{\mathop{\mathrm{l}}}
\newcommand{\sgn}{\mathop{\mathrm{sign}}}
\newcommand{\sym}{\mathop{\mathrm{sym}}}
\newcommand{\avg}{\mathop{\mathrm{avg}}}
\newcommand{\mean}{\mathop{\mathrm{mean}}}
\newcommand{\erf}{\mathop{\mathrm{erf}}}
\newcommand{\grad}{\nabla}
\newcommand{\R}{\mathbb{R}}
\newcommand{\defeq}{\triangleq}
\newcommand{\dims}[2]{[#1\!\times\!#2]}
\newcommand{\sdims}[2]{\mathsmaller{#1\!\times\!#2}}
\newcommand{\udims}[3]{#1}
\newcommand{\udimst}[4]{#1}
\newcommand{\com}[1]{\rhd\text{\emph{#1}}}
\newcommand{\ind}{\hspace{1em}}
\newcommand{\argmin}[1]{\underset{#1}{\operatorname{argmin}}}
\newcommand{\floor}[1]{\left\lfloor{#1}\right\rfloor}
\newcommand{\step}[1]{\vspace{0.5em}\noindent{#1}}
\newcommand{\quat}[1]{\ensuremath{\mathring{\mathbf{#1}}}}
\newcommand{\norm}[1]{\left\lVert#1\right\rVert}
\newcommand{\ignore}[1]{}
\newcommand{\specialcell}[2][c]{\begin{tabular}[#1]{@{}c@{}}#2\end{tabular}}
\newcommand*\Let[2]{\State #1 $\gets$ #2}
\newcommand{\algorithmicbreak}{\textbf{break}}
\newcommand{\Break}{\State \algorithmicbreak}
\newcommand{\ra}[1]{\renewcommand{\arraystretch}{#1}}

\renewcommand{\vec}[1]{\mathbf{#1}} 

\algdef{S}[FOR]{ForEach}[1]{\algorithmicforeach\ #1\ \algorithmicdo}
\algnewcommand\algorithmicforeach{\textbf{for each}}
\algrenewcommand\algorithmicrequire{\textbf{Require:}}
\algrenewcommand\algorithmicensure{\textbf{Ensure:}}
\algnewcommand\algorithmicinput{\textbf{Input:}}
\algnewcommand\INPUT{\item[\algorithmicinput]}
\algnewcommand\algorithmicoutput{\textbf{Output:}}
\algnewcommand\OUTPUT{\item[\algorithmicoutput]}

\maketitle
\thispagestyle{empty}
\pagestyle{empty}

\begin{abstract}
Affordance detection is a challenging problem with a wide variety of robotic applications. Traditional affordance detection methods are limited to a predefined set of affordance labels, hence potentially restricting the adaptability of intelligent robots in complex and dynamic environments. In this paper, we present the Open-Vocabulary Affordance Detection (OpenAD) method, which is capable of detecting an unbounded number of affordances in 3D point clouds. By simultaneously learning the affordance text and the point feature, OpenAD successfully exploits the semantic relationships between affordances. Therefore, our proposed method enables zero-shot detection and can be able to detect previously unseen affordances without a single annotation example. Intensive experimental results show that OpenAD works effectively on a wide range of affordance detection setups and outperforms other baselines by a large margin. Additionally, we demonstrate the practicality of the proposed OpenAD in real-world robotic applications with a fast inference speed. Our project is available at \href{https://openad2023.github.io}{https://openad2023.github.io}.

\end{abstract}


\section{INTRODUCTION} \label{Sec:Intro}

The concept of affordance, proposed by the ecological psychologist James Gibson~\cite{gibson1966senses}, plays an important role in various robotic applications, such as object recognition~\cite{thermos2017deep,hou2021affordance}, action anticipation~\cite{jain2016structural,roy2021action}, agent's activity recognition~\cite{vu2014predicting,qi2017predicting,chen2023affordance}, and object functionality understanding~\cite{li2023locate,jiang2022a4t}. 
In these applications, affordances are used to illustrate the potential interactions between the robot and its surrounding environment.
For instance, with a general cutting task, the knife's affordances can guide the robot to use the knife's blade to achieve requirements such as mincing meat or carving wood. Detecting object affordances, however, is not a trivial task since the robots need to understand in real-time the arbitrary correlations between objects, actions, and effects in complex and dynamic environments \cite{min2016affordance}.

Traditional methods for affordance detection utilize classical machine learning methods on the images, such as Support Vector Machine (SVM) based affordance prediction~\cite{hermans2011affordance}, texture-based and object-level monocular appearance cues~\cite{song2011visual}, relational affordance model~\cite{moldovan2014occluded}, and human-object interactions~\cite{hassan2016attribute}. With the rise of deep learning, several works have employed Convolutional Neural Networks (CNN)~\cite{krizhevsky2017imagenet} for different affordance-related tasks, such as affordance reasoning~\cite{mottaghi2017see,chuang2018learning}, pixel-based affordance detection~\cite{nguyen2016detecting,do2018affordancenet,luo2023leverage,nguyen2017object}, and functional scene understanding~\cite{li2019putting,pacheco2023one}.
The key challenge in detecting object affordances via the imagery data is that object affordances may differ in terms of visual information such as shape, size, or geometry while being similar in object functionality~\cite{hassanin2021visual}.
In practice, object affordance detection from the images requires an additional step to be applied to downstream robotic tasks, as we need to transform the detected results from 2D to 3D using the depth information~\cite{deng20213d}.

With the increasing availability of advanced depth cameras, 3D point cloud has become a popular modality for robotic applications~\cite{liu2019deep}. 
Compared to conventional images, 3D point clouds directly provide the robot with 3D information about surrounding objects and the environment. 
Consequently, several recent works directly utilize the 3D point clouds for affordance detection~\cite{kokic2017affordance,deng20213d,iriondo2021affordance,mo2022o2o}.
For instance, Kim~\etal~\cite{kim2014semantic} detected affordances by dividing point clouds into segments and classifying them using logistic regression. The authors in~\cite{ten2017grasp} proposed a new grasp detection method in point cloud data. 
More recently, Mo~\etal~\cite{deng20213d} predicted affordance heat maps from human-object interaction via the scene point cloud.
In this work, we address the task of affordance detection in 3D point clouds to directly apply the results to robotic application tasks.
More specifically, we consider the affordances of point-level objects and propose a new method to generalize the affordance understanding using the open-vocabulary setting.

\begin{figure}
\centering
\subfigure[]{\label{fig:sub_intro1}\includegraphics[height=50mm]{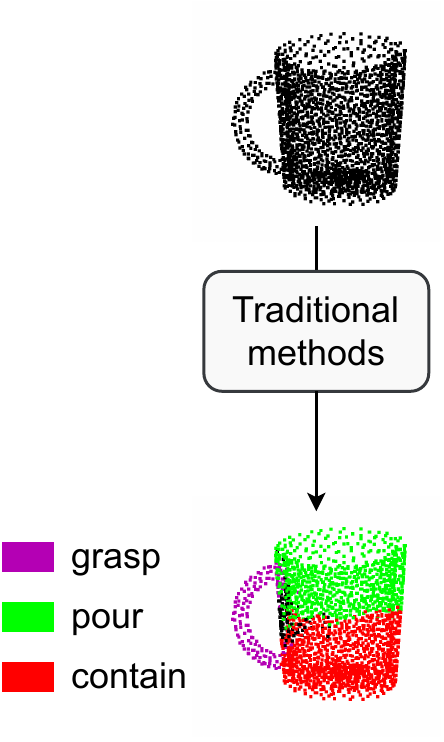}}
\subfigure[]{\label{fig:sub_intro2}\includegraphics[height=50mm]{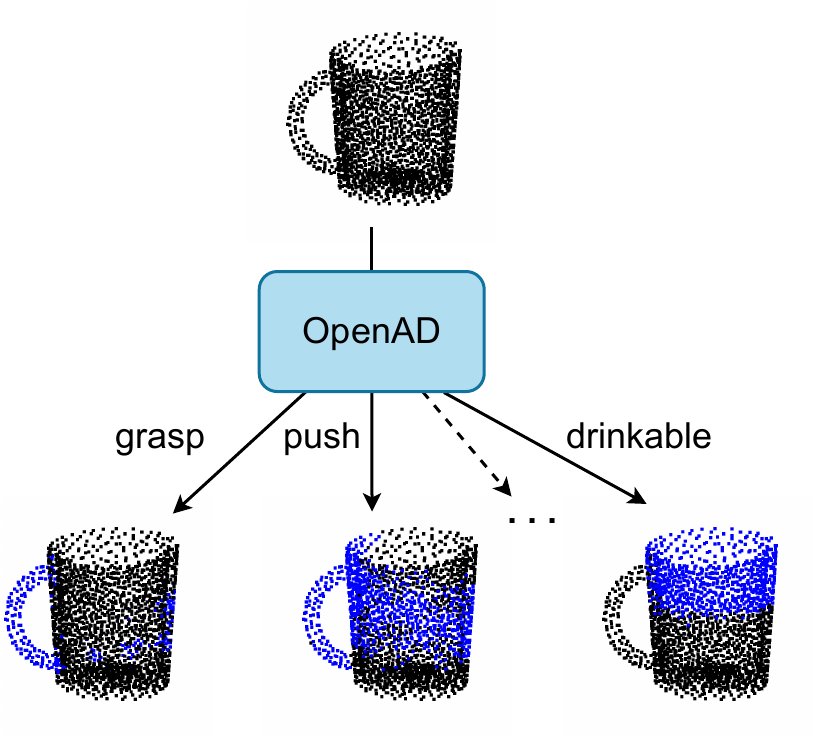}}
\vspace{0.5ex}
\caption{The comparison between traditional affordance detection methods (a) and our method (b). Traditional methods are restricted to predefined affordance label sets, while our OpenAD enables open-set affordance labels.}
\label{fig:intro}
\end{figure}

While several works have been proposed for affordance understanding using 2D images or 3D point clouds, they are mostly restricted to a~\emph{predefined affordance label set}. 
This limitation prevents robots from quickly adapting to a wide range of real-world scenarios or responding to changes in the operating environments. 
Recently, increasing the flexibility of affordance labels has been studied as a one-shot or few-shot learning problem~\cite{luo2021one,zhai2022one}.
However, they only consider the 2D images as input and treat the problem as the classical pixel-wise affordance segmentation task.
In this work, we overcome the limitation of the fixed affordance label set by addressing the affordance detection in 3D point clouds under the~\emph{open-vocabulary} setting. Our key idea is to learn collaboratively the mapping between the \emph{language labels} and the \emph{visual features} of the point cloud. In contrast to traditional methods that are limited to a predefined set of affordance labels, our approach allows the robot to utilize an unrestricted number of natural language texts as input and, therefore, can be used in a broader range of applications. The main concept of our approach is illustrated in Figure~\ref{fig:intro}. Unlike~\cite{zhai2022one}, our method does not require annotation examples for unseen affordances and can also work directly with 3D data instead of 2D images.

In this paper, we present
Open-Vocabulary Affordance Detection (OpenAD).
Our main goal is to provide a framework that does not restrict the application to a fixed affordance label set.
Our method takes advantage of the recent large-scale language models, i.e., CLIP~\cite{radford2021learning}, and enhances the generalizability of the affordance detection task in 3D point clouds.
Particularly, we propose a simple, yet effective method for learning the affordance label and the visual feature together in a shared space.
Our method enables zero-shot learning with the ability to process new open-language texts as query affordances.

Our contributions are summarized as follows:
\begin{itemize}
    \item
    We present OpenAD, a simple but effective method to tackle the task of open-vocabulary affordance detection.
    \item
    We conduct intensive experiments to validate our method and demonstrate the usability of OpenAD in real-world robotic applications.
\end{itemize}

\section{Related Work} \label{Sec:rw}
\textbf{Pixel-Wise Affordance Detection.} A large number of works consider affordance detection as a pixel-wise labeling task, see, e.g., \cite{nguyen2016detecting,roy2016multi,nguyen2017object,do2018affordancenet,thermos2020deep,chen2022cerberus,luo2022learning}.
Nguyen \etal \cite{nguyen2017object} detected object affordances in real-world scenes using an object detector and dense conditional random fields (CRF). 
The authors in~\cite{do2018affordancenet} proposed a two-branch framework that simultaneously recognizes multiple objects and their affordances from RGB images. 
Chen~\etal~\cite{chen2022cerberus} presented a multi-task dense prediction architecture and a tailored training framework to address the problem of joint semantic, affordance, and attribute parsing. 
More recently, Luo~\etal~\cite{luo2022learning} proposed a cross-view knowledge transfer framework to extract invariant affordances from exocentric observations and transfer them to egocentric views. 
Some other works designed different settings for their task of affordance detection on images~\cite{hassan2016attribute,chen2015deepdriving}. 
In particular, Hassan~\etal~\cite{hassan2016attribute} predicted high-level affordances by investigating mutual contexts of humans, objects, and the ambient environment, while Chen~\etal~\cite{chen2015deepdriving} learned meaningful affordance indicators to predict actions for autonomous driving.

\textbf{Affordance Detection in 3D Point Clouds.} Affordance detection is also studied on 3D point cloud data~\cite{kim2014semantic,kim2015interactive,kokic2017affordance,deng20213d,iriondo2021affordance,mo2022o2o}. 
Particularly, Kim~\etal~\cite{kim2014semantic} presented a method to extract geometric features from point cloud segments and classify affordances by logistic regression.
Subsequently, Kim and Sukhame~\cite{kim2015interactive} presented a technique that voxelizes point cloud objects and creates affordance maps through interactive manipulation.
Kokic~\etal~\cite{kokic2017affordance} proposed a system for modeling relationships between task, object, and a grasp to address the problem of task-specific robot grasping. 
Also focusing on detecting affordances for the task of grasping, the authors in~\cite{iriondo2021affordance} localized grasping affordances for industrial bin-picking applications. 
Lately, Mo~\etal~\cite{mo2022o2o} learned affordance heatmaps from object-object interaction. 
Regardless of performing on images or 3D point clouds, none of the above methods addressed the task of open-vocabulary affordance detection. 

\textbf{Language-Driven Segmentation.} 
Language-driven segmentation has recently attracted research interest in computer vision and machine learning.
Promising results of language-driven segmentation are achieved by large-scale language models such as CLIP~\cite{radford2021learning} or BERT~\cite{devlin2018bert}. 
Inspired by these works, several researchers attempted to apply the same idea to other domains, including semantic segmentation. 
For instance, Li~\etal~\cite{li2021language} aligned pixel-wise features and text features to tackle the task of zero-shot semantic segmentation. 
With a similar focus, Xu~\etal~\cite{xu2022simple} proposed a two-stage framework that first extracts mask proposals and then performs classification on the masked image crops generated. 
Rozenberszki~\etal~\cite{rozenberszki2022language} proposed a language-driven contrastive pre-training method to address the task of indoor semantic segmentation. 
Using the pre-trained text encoder from CLIP for open-vocabulary 3D scene understanding, Peng~\etal~\cite{peng2022openscene} proposed a technique that fused 2D and 3D features and aligned the combination with text embeddings. 
Despite recent development in the field of language-driven segmentation, most recent works have focused only on 2D images, while the task of open-vocabulary affordance detection in 3D point clouds remains an open problem.
 
\section{Open-Vocabulary Affordance Detection} \label{Sec:method}

\begin{figure*}[t]
	\centering
	\includegraphics[width=1.0\linewidth]{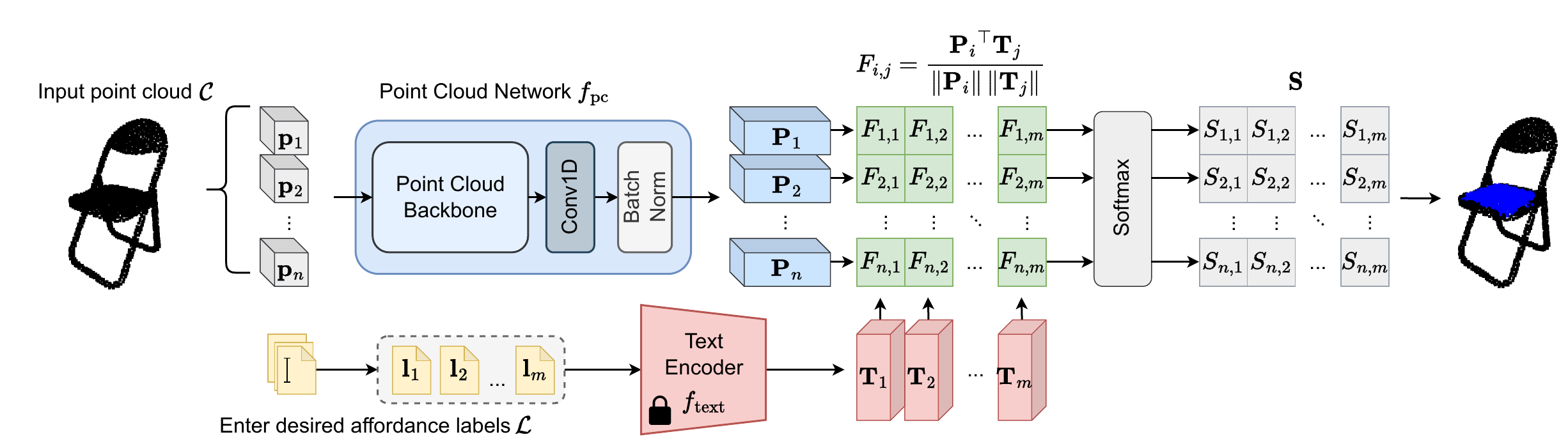}
 \vspace{1pt}
	\caption{The overview of the proposed Open-Vocabulary Affordance Detection (OpenAD) network. 
                 First, the input point cloud is fed into a point cloud network to extract per-point embeddings. 
                 Second, the affordance labels are passed into a text encoder to extract the text embeddings.
                 Subsequently, the correlation between the point-wise features and the corresponding text embeddings is computed using the cosine similarity function. 
                 Finally, a softmax layer is employed to predict language-driven affordances.}
	\label{fig:architecture}
\end{figure*}

\subsection{Problem Formulation}
We consider the task of open-vocabulary affordance detection in 3D point clouds by training a combined vision-language model. 
Specially, we take into account an input point cloud $\mathcal{C} = \left \{\mathbf{p}_1, \mathbf{p}_2,...,\mathbf{p}_n  \right \}$ of $n$ unordered points, $\mathbf{p}_i \in \mathbb{R}^3$, $i=1,...,n$. 
Each point is represented by its coordinate in Euclidean space. 
The affordance labels are presented in a natural language form, i.e., $\mathcal{L} = \left \{\mathbf{l}_1, \mathbf{l}_2,..., \mathbf{l}_m \right \}$. Note that, in our open-vocabulary label setting, the number of labels $m$ can ideally be unlimited. The testing label set can differ from the training label set and can contain unseen affordance labels.
Our goal is to jointly learn the affordance labels and the input point cloud. 
We first extract the point cloud feature using a deep point cloud network. 
The affordance label is encoded by a text encoder. 
We then introduce a simple, however, effective correlation metric to jointly learn the point cloud visual feature and the text embedding features. 
In this manner, our method leverages the similarities of text embeddings of unseen affordances that are semantically related to the ones seen in the training process.
The overall framework of our method is depicted in Figure~\ref{fig:architecture}.
\subsection{Open-Vocabulary Affordance Detection}
\textbf{Text encoder.}~The text encoder $f_{\rm{text}}\left ( \cdot\right )$ embeds the set of potential affordance labels into an embedding space $\mathbb{R}^D$. 
Similar to other works~\cite{li2021language,xu2022simple}, the text encoder can be an arbitrary network.
In this work, we employ the state-of-the-art pre-trained ViT-B/32 text encoder from CLIP~\cite{radford2021learning}. 
The text encoder produces $m$ word embeddings $\mathbf{T}_1, \mathbf{T}_2,...,\mathbf{T}_m \in\mathbb{R}^D$ as the presentation of the input affordance labels.
Note that we only use the CLIP text encoder to extract the text features and freeze $f_{\rm{text}}\left ( \cdot\right )$ during both training and testing.

\textbf{Point cloud network.}~The second component of our method is the point cloud network. The $n$ input points are plugged into the point cloud network $f_{\rm pc}\left ( \cdot\right )$ producing an embedding vector for every input point. Similar to the text encoder, the architecture of the point cloud network can be various. 
In this work, we use the state-of-the-art point cloud model, i.e., PointNet++~\cite{qi2017pointnet++} as the underlying architecture. 
Furthermore, we append to the end of the backbone with a convolutional layer with $D$ output units shared across all points, followed by a batch norm layer. 
More specifically, the point cloud network $f_{\rm pc}\left ( \cdot\right )$ produces a set of $n$ vectors $\mathbf{P}_1, \mathbf{P}_2,...,\mathbf{P}_n\in\mathbb{R}^D$. 
In contrast to the text encoder, the weights of the point cloud network $f_{\rm pc}\left ( \cdot\right )$ are updated during the training.

\textbf{Learning text-point correlation.}
To enable open-vocabulary affordance detection, the semantic relationships between the point cloud affordance and its potential labels have to be computed. 
Particularly, we correlate point-wise embeddings of the input point cloud $\mathbf{P}_{i}$ and text embeddings of affordance labels $\mathbf{T}_{j}$ using the cosine similarity function. 
The correlation value $F_{i,j}$, which is an element of the $i$-th row and $j$-th column of the correlation matrix $\mathbf{F} \in \mathbb{R}^{n\times m}$, is computed as
\begin{equation}
{F}_{i,j} = \frac{{\mathbf{P}^\top_{i}} \mathbf{T}_j}{\left \| \mathbf{P}_i \right \|\left \| \mathbf{T}_j \right \|}\:\:.
\end{equation}
The point-wise softmax output of a single point $i$ is then computed in the form
\begin{equation}
    {S}_{i,j}= 
    \dfrac{\exp\left ({F}_{i,j}/\tau\right )}
    {\sum_{k=1}^{m}\exp({{F}_{i,k}}/\tau)}\:\:,
\label{eq: S}    
\end{equation}
where $\tau$ is a learnable parameter~\cite{wu2018unsupervised}. This computation is applied for every point in the point cloud.
During the training, we encourage the point cloud network $f_{\rm pc} \left (\cdot \right )$ to provide point embeddings that are close to the text embeddings. 
These text embeddings are produced by the text encoder $f_{\rm text}\left (\cdot \right )$ of the corresponding ground-truth classes. 
Specifically, given the embedding $\mathbf{P}_i\in\mathbb{R}^D$ of point $i$, we aim to maximize the value of the entry ${F}_{i,j}$ that is the similarity of $\mathbf{P}_i$ and the text embedding $\mathbf{T}_j$ corresponding to the ground-truth label $j = y_i$.
This can be accomplished by optimizing the weighted negative log-likelihood loss of the point-wise softmax output over the entire point cloud in the form
\begin{equation}
    {L = -\sum_{i=1}^n w_{y_i}\log {S}_{i,{y_i}}} \:\:,
\end{equation}
where $w_{y_i}$ is the weighting parameter to the imbalance problem of the label classes during the training. 
Inspired by~\cite{qi2017pointnet++}, we define this weight as
\begin{equation}
    w_j = \left(\frac{\mathrm{max}\left \{c_1,c_2,...,c_{m} \right \}}{c_j}\right)^{1/3},
\end{equation}
where $c_j$ is the number of points of the class $j$ of the training set.

\subsection{Training and Inference}
During the training, we fix the text encoder and train the rest of the network end-to-end. 
Similar to~\cite{deng20213d}, we fix the number of points in a point cloud to $n = 2048$. We set $D$ to $512$.
We train our network using the Adam optimizer with the learning rate $\alpha = 10^{-3}$ and the weight decay $\gamma = 10^{-4}$. 
The proposed framework is trained over $200$ epochs on a 24GB-RAM NVIDIA Geforce RTX 3090 Ti with a batch size of $16$. 
We initialize the value of $\ln{(1/0.07)}$ for $\tau$. 
During the inference, we feed the point cloud and any text as the input label to detect the wanted affordance from the point cloud. 
The inference process takes approximately \SI{100}{\milli\second} on average with our proposed network.

\section{Experiments} \label{Sec:exp}
In this section, we perform several experiments to validate the effectiveness of our OpenAD. We start with a zero-shot detection setting to verify the ability of OpenAD to generalize to previously unseen affordances. Secondly, we present OpenAD's notable qualitative results together with visualizations. Finally, we conduct additional ablation studies to further investigate other aspects of OpenAD.

\subsection{Zero-Shot Open-Vocabulary Affordance Detection}

\textbf{Dataset.}
We use the 3D AffordanceNet dataset~\cite{deng20213d} in our experiments. 3D AffordanceNet dataset is currently the largest dataset for affordance understanding using 3D point cloud data with $22,949$ instances from $23$ object categories and $18$ affordance labels. As in other zero-shot setups~\cite{cheraghian2019zero,cheraghian2020transductive,michele2021generative}, we need more label classes to verify the robustness of the methods, therefore, we re-label the 3D AffordanceNet with the extra $18$ affordance classes and also consider the background as a class, hence making the total number of affordance labels $37$. Following~\cite{deng20213d}, we benchmark on the two tasks: the full-shape and partial-view tasks. 
The partial-view setup is more useful in robotics as the robot usually can only observe a partial-view of the object's point cloud.

\textbf{Baselines and Evaluation Metrics.}
We compare our method with the following recent methods for zero-shot learning in 3D point clouds: ZSLPC~\cite{cheraghian2019zero}, TZSLPC~\cite{cheraghian2020transductive}, and 3DGenZ~\cite{michele2021generative}. Note that, since these baselines used GloVe~\cite{pennington2014glove} or Word2Vec~\cite{mikolov2013distributed} for word embedding, which are less powerful models than CLIP, we replace their original text encoders with CLIP for a fair comparison. For ZSLPC~\cite{cheraghian2019zero} and TZSLPC~\cite{cheraghian2020transductive}, we change their classification heads to the segmentation task. As in~\cite{qi2017pointnet++,wang2019dynamic,zhao2021point}, we use three metrics to evaluate the results: mIoU (mean IoU over all classes), Acc (overall accuracy over all points), and mAcc (mean accuracy over all classes). 

\textbf{Resutls.} Table~\ref{tab: zero-shot task result} shows that our OpenAD achieves the best results on both tasks and all three metrics. In particular, on the full-shape task, OpenAD significantly surpasses the runner-up model (ZSLPC) by 4.40\% on mIoU. OpenAD also outperforms others on Acc (0.84\% over 3DGenZ) and on mAcc (0.81\% over ZSLPC). Similarly, for the partial-view task, our method has the highest mIoU (3.98\% higher than the second-best ZSLPC), and also the highest Acc and mAcc.

\begin{table}[ht]
\caption{Zero-shot Open-vocabulary detection results}
\label{tab: zero-shot task result}
\vskip 0.15in
\begin{center}
\begin{tabular}{llccc}
\toprule
Task & Method  &  mIoU & Acc & mAcc \\
\midrule
Full-shape & TZSLPC\cite{cheraghian2020transductive} & 3.86 & 42.97 & 10.37 \\
& 3DGenZ~\cite{michele2021generative} & 6.46 & 45.47 & 18.33 \\
& ZSLPC~\cite{cheraghian2019zero} & 9.97 & 40.13 & 18.70 \\
& OpenAD (ours) & \bf{14.37} & \bf{46.31} & \bf{19.51} \\
\midrule
Partial-view & TZSLPC~\cite{cheraghian2020transductive} & 4.14 & 42.76 & 8.49 \\
& 3DGenZ~\cite{michele2021generative} & 6.03 & 45.24 & 15.86 \\
& ZSLPC~\cite{cheraghian2019zero} & 9.52 & 40.91 & 17.16 \\
& OpenAD (ours) & \bf{12.50} & \bf{45.25} & \bf{17.37} \\
\bottomrule
\end{tabular}
\end{center}
\vskip -0.1in
\end{table}

\subsection{Qualitative Results}
We present several examples to demonstrate the generality and flexibility of OpenAD. Primarily, we use objects from the 3D AffordanceNet~\cite{deng20213d} for our visualizations. We also select objects from the ShapeNetCore dataset~\cite{chang2015shapenet} to analyze the capability of OpenAD to generalize to unseen object categories and new affordance labels.

\begin{figure*}[t]
	\centering
	\includegraphics[width=0.88\linewidth]{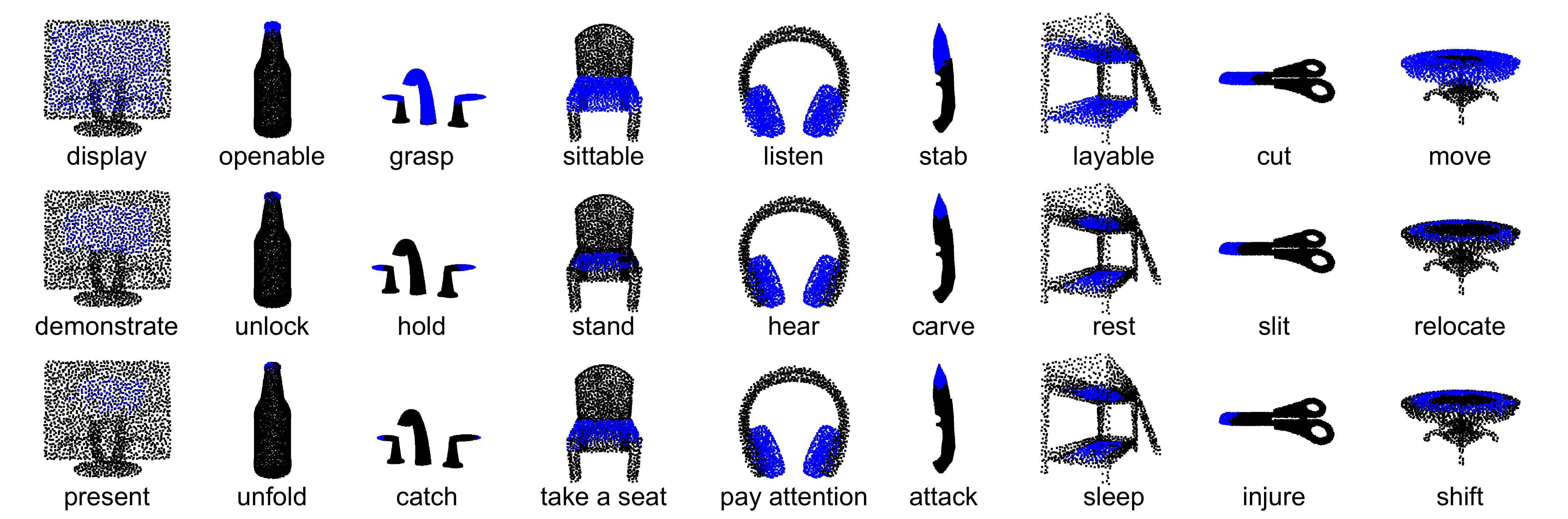}
 \vspace{1.2ex}
	\caption{The visualization of our OpenAD's capability to detect seen and unseen affordances. \textit{The upper row:} Detection results of OpenAD on seen affordances. \textit{The middle and lower rows:} Detection results of OpenAD on unseen affordances that do not exist in the training set.}
	\label{fig:unseen_affordances}
\end{figure*}

\textbf{Generalization to New Affordance Labels.}
We illustrate several examples showing the ability of OpenAD to generalize to unseen affordance classes in Figure~\ref{fig:unseen_affordances}. In the upper row of Figure~\ref{fig:unseen_affordances}, we present the detection results of OpenAD for nine seen affordances on appropriate objects. As trained on these affordances, OpenAD produces good detection results. Next, for each object, we feed new affordance labels to the models while keeping the same corresponding object, and present the detection results in the two rows below. The visualization shows that OpenAD successfully detects the associated regions for the queried new affordance labels, even though the labels are not included in the training set.

\begin{figure}[t]
	\centering
	\includegraphics[width=0.9\linewidth]{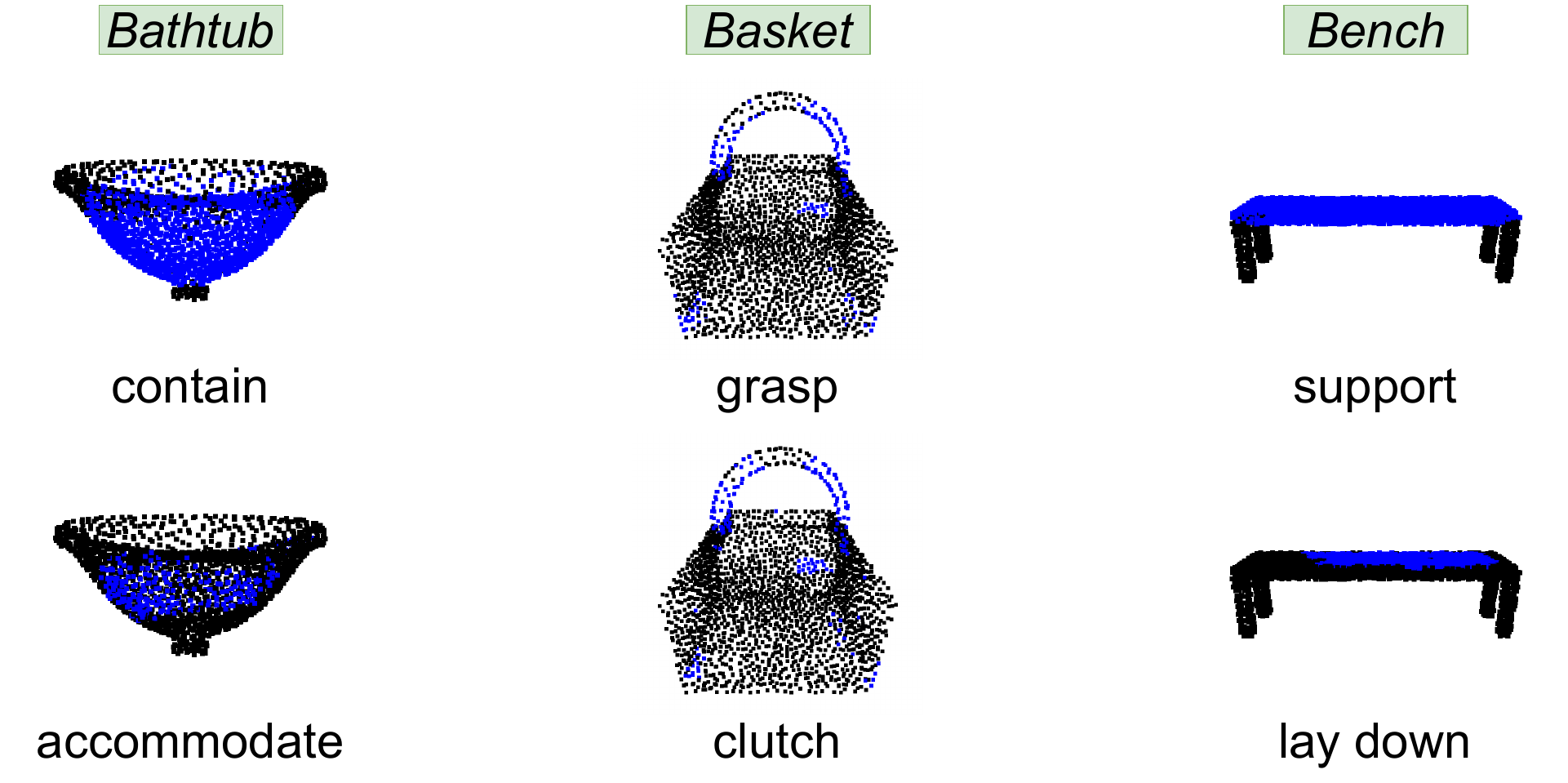}
 \vspace{1pt}
	\caption{Results on unseen object categories. OpenAD gives reasonable outputs when detecting affordances on object categories that are not in the training set. \textit{Upper row:} Seen affordances. \textit{Lower row:} Unseen affordances.}
	\label{fig:unseen_objects}
\end{figure}

\textbf{Generalization to Unseen Object Categories.}
In this work, OpenAD is trained on 3D AffordanceNet dataset~\cite{deng20213d}, which covers 23 object categories. To verify the generalization ability of our method on new unseen objects, we select new novel objects from the ShapeNetCore dataset~\cite{chang2015shapenet} and test them in both cases: seen and open-vocabulary affordance labels. We use the farthest point sampling algorithm to uniformly sample $2,048$ points from the surface of each object. All points are then centered and scaled before being fed into OpenAD.
Figure~\ref{fig:unseen_objects} summarises the results. This figure shows that OpenAD is able to detect the affordance classes on new object categories. This confirms the generalization of our OpenAD for downstream robotic tasks.

\begin{figure}[ht]
	\centering
	\includegraphics[width=0.8\linewidth]{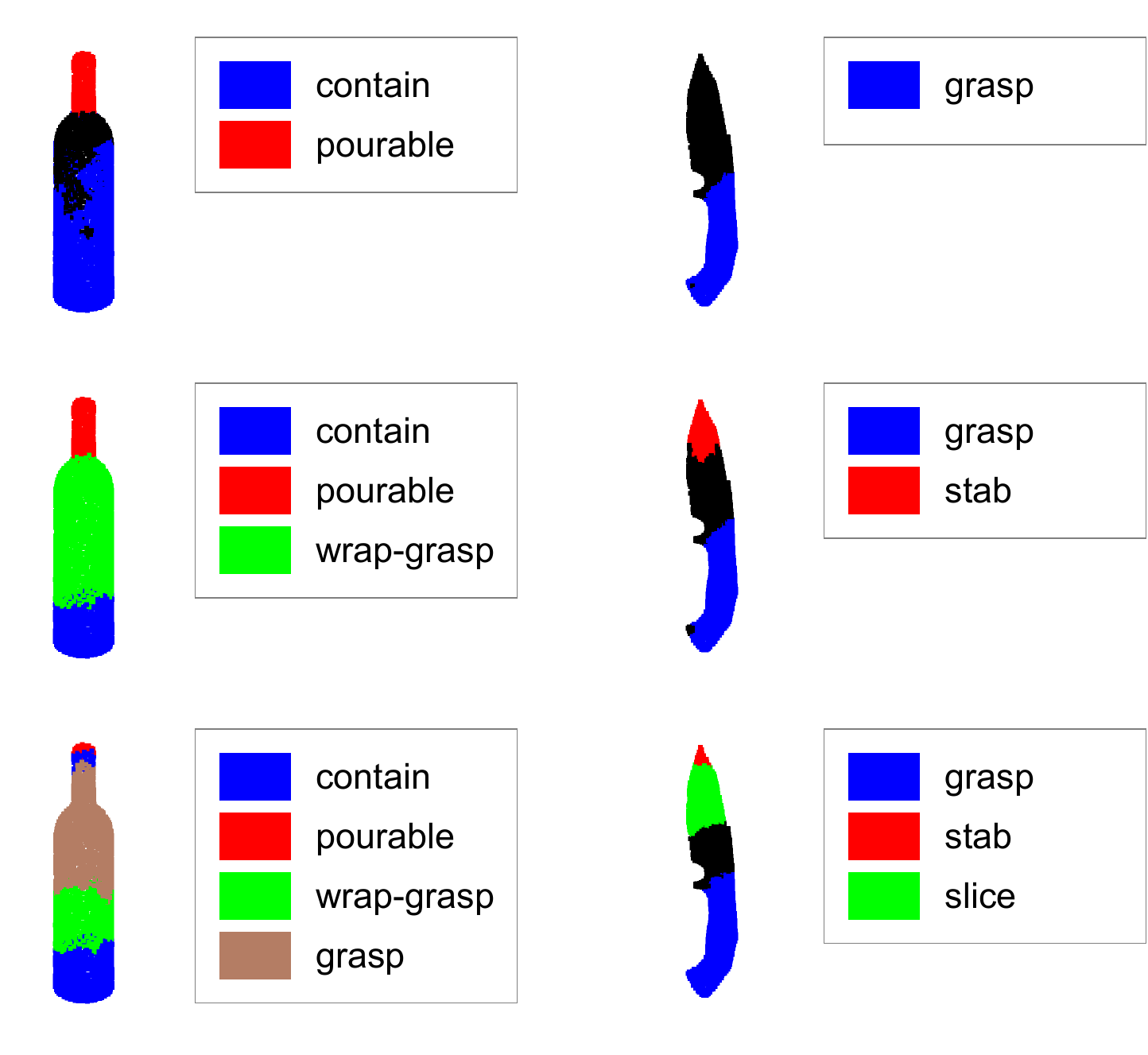}
 \vspace{1ex}
	\caption{Multi-affordance detection using our method.
 }
	\label{fig:multilabel}
\end{figure}

\textbf{Multi-Affordance Detection.}
In OpenAD, the number of affordances in the label set, $m$, can vary. This flexible design allows OpenAD to detect multiple affordances at once. Figure~\ref{fig:multilabel} presents the detection results of two objects given label sets with different numbers of affordances. Moreover, we observe a notable ability of OpenAD that it does not fix the label for a particular point but finds the most suitable label in a specific label set. Concretely, given an object, OpenAD first detects affordances in a particular label set.
By maintaining the earlier affordances, once a new affordance label is added to the label set, there are certain points will be labeled by new affordances over the previous affordance classes.
For instance, in the left column of Figure~\ref{fig:multilabel}, the points in the upper body of the bottle are labeled as~\texttt{contain} in the first run, then they are re-labeled as \texttt{wrap-grasp} in the second run and finally as \texttt{grasp} in the last run. This OpenAD's ability can also be observed in the case of the knife object in the right column of Figure~\ref{fig:multilabel}. It further demonstrates our OpenAD's flexibility.

\subsection{Ablation Study}

\textbf{Will language-driven architecture affect the detection results?} In this work, we mainly focus on jointly learning a vision-language model to improve the generalization in downstream robotic tasks, and do not aim for improving the accuracy of the traditional affordance detection tasks. This leads to the question that whether our language-driven model affects the accuracy of the traditional affordance detection task. To verify this, we train our OpenAD on the original 3D AffordanceNet with its original label set and training split, and compare our result with other state-of-the-art methods, including PointNet++~\cite{qi2017pointnet++}, Dynamic Graph CNN (DGCNN)~\cite{wang2019dynamic} and Point Transformer~\cite{zhao2021point}. For PointNet++ and DGCNN, we follow similar designs in~\cite{deng20213d} and change the final classifier to a linear layer detecting the affordance classes. For Point Transformer, we apply the same architecture in~\cite{zhao2021point}. Table~\ref{tab: original task result} presents the results of all methods on 3D AffordanceNet~\cite{deng20213d}. From this table, we find that OpenAD performs competitively when compared to other methods. Therefore, we can conclude that while our method is designed for a different purpose, it still can be used as a strong benchmark for closed-set affordance detection.

\begin{table}[t]
\caption{Closed-set detection results}
\label{tab: original task result}
\vskip 0.15in
\begin{center}
\begin{tabular}{llccc}
\toprule
Task & Method  &  mIoU & Acc & mAcc \\
\midrule
Full-shape & {Point Transformer}~\cite{zhao2021point} & 41.26 & 68.67 & 67.03 \\
& {PointNet++}~\cite{qi2017pointnet++} & 41.26 & 68.51 & \textbf{68.14} \\
& DGCNN~\cite{wang2019dynamic} & \bf{42.09} & 68.47 & 61.47 \\
& {OpenAD} (ours) & {42.00} & \bf{69.03} & 67.31 \\
\midrule
Partial-view & {Point Transformer}~\cite{zhao2021point} & 40.51 & 69.22 & 65.34 \\
& {PointNet++}~\cite{qi2017pointnet++} & 41.10 & 69.40 & \bf{66.74} \\
& DGCNN~\cite{wang2019dynamic} & \bf{41.93} & 69.38 & 63.12 \\
& {OpenAD} (ours) & {41.87} & \bf{69.91} & 66.34 \\
\bottomrule
\end{tabular}
\end{center}
\vskip -0.1in
\end{table}

\textbf{Backbones and Text Encoders.} In table~\ref{tab: ablation study}, we conduct an ablation study on two different point cloud backbones, i.e., PointNet++~\cite{qi2017pointnet++} and DGCNN~\cite{wang2019dynamic}, and two different pre-trained text encoders, i.e., CLIP ViT-B/32~\cite{radford2021learning} and BERT~\cite{devlin2018bert}. Note that with BERT, the parameter $D$ is set to $768$. We observe that different combinations of backbones and text encoders perform equivalently on the closed-set tasks. Meanwhile, on the open-vocabulary tasks, PointNet++ performs better than DGCNN, and CLIP performs better than BERT. The gap in the performance of frameworks using CLIP ViT-B/32 text encoder compared to those using BERT is significant, demonstrating the superiority of CLIP in semantic language-vision understanding.

\begin{table}[ht]
\caption{Ablation study on point cloud backbone and text encoder}
\label{tab: ablation study}
\vskip 0.15in
\begin{center}
\begin{tabular}{lllccc}
\toprule
\makecell[l]{Task \&\\setting} & \makecell[l]{Point cloud\\backbone} & \makecell[l]{Text \\ encoder} & mIoU & Acc & mAcc \\
\midrule
{Full-shape \&} & PointNet++ & CLIP & \bf{14.37} & 46.31 & \bf{19.51} \\
Open-vocabulary &  PointNet++ & BERT & 10.55 & \bf{46.81} & 16.02 \\
&  DGCNN & CLIP & 10.88 & 45.21 & 15.40 \\
&  DGCNN & BERT & 9.43 & 46.37 & 14.61 \\
\midrule
{Partial-view \&} & PointNet++ & CLIP & \bf{12.50} & 45.25 & \bf{17.37} \\
Open-vocabulary &  PointNet++ & BERT & 9.33 & \bf{45.71} & 14.45 \\
&  DGCNN & CLIP & 11.19 & 44.21 & 16.43 \\
&  DGCNN & BERT & 7.00 & 44.93 & 10.97 \\
\midrule
{Full-shape \&} & PointNet++ & CLIP & 42.00 & \bf{69.03} & \bf{67.31} \\
Closed-set &  PointNet++ & BERT & 41.04 & 68.79 & 67.22 \\
&  DGCNN & CLIP & \bf{42.11} & 68.59 & 61.31 \\
&  DGCNN & BERT & 42.06 & 68.31 & 61.26 \\
\midrule
{Partial-view \&} & PointNet++ & CLIP & \bf{41.87} & \bf{69.91} & \bf{66.34} \\
Closed-set &  PointNet++ & BERT & 41.50 & 69.77 & 65.55 \\
&  DGCNN & CLIP & 41.20 & 69.16 & 64.16 \\
&  DGCNN & BERT & 41.27 & 69.62 & 58.09 \\
\bottomrule
\end{tabular}
\end{center}
\vskip -0.1in
\end{table}

\subsection{Robotic Demonstration}
\begin{figure}
\centering
\subfigure{
\label{fig:rotbot1}
\def\svgwidth{1\columnwidth}
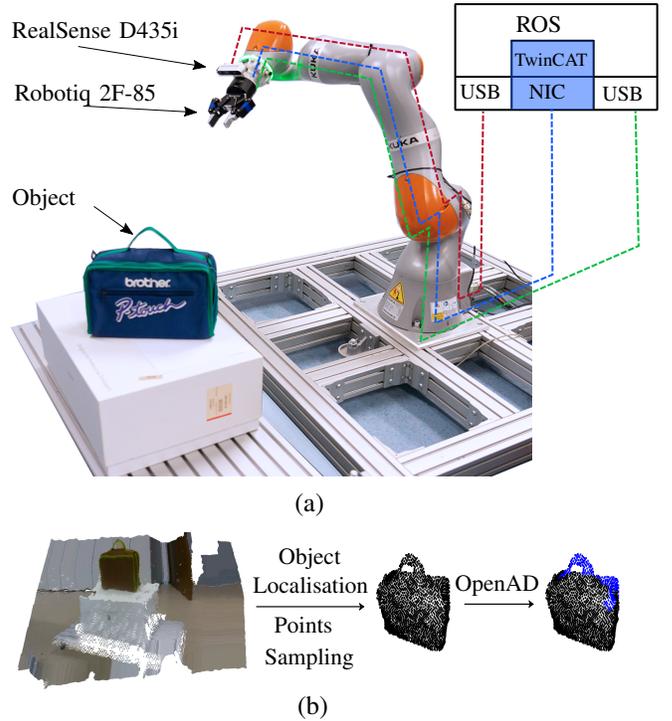
}
\vspace{1pt}
\caption{Example of a robot demonstration. (a) Experimental setup. (b) Result from OpenAD.}
\label{fig: robot demonstration}
\end{figure}

The experiment setup, shown in Figure~\ref{fig: robot demonstration}, comprises five main components, i.e. the robot \kuka, the PC1 running the real-time automation software Beckhoff TwinCAT, the Intel RealSense D435i camera, the Robotiq 2F-85 gripper, and the PC2 running Robot Operating System (ROS) Noetic 20.04. 
PC1 communicates with the robot via a network interface card (NIC) using the EtherCAT protocol, marked by the blue region in Figure~\ref{fig: robot demonstration}. Note that the robot control is implemented in a C++ module in PC1. The sampling time is set to $\SI{125}{\micro\second}$ for the robot sensors and actuators. 
PC2 controls the gripper and the camera via the USB protocol in the ROS environment. 
Additionally, the two PCs communicate with each other via an Ethernet connection. After receiving point cloud data of the environment from the RealSense D435i camera, we utilize the state-of-the-art object localization method~\cite{zhou2018voxelnet} to identify the object, then perform point sampling to get $2048$ points.  
We then feed this point cloud with a natural language affordance command.
Note that, using our OpenAD, we can have a general input command and are not restricted to a predefined affordance label set. 
Our OpenAD returns the affordance region, which can be used for the grasp pose detection module~\cite{ten2017grasp}, the analytical inverse kinematics module~\cite{vu2022machine}, and the trajectory optimization module~\cite{beck2022singlularity}.
Several demonstrations, such as holding and raising a bag, wrapping a bottle, and pushing an earphone, can be found in our supplementary material.

\subsection{Discussion}
\begin{figure}[t]
	\centering
	\includegraphics[width=0.95\linewidth]{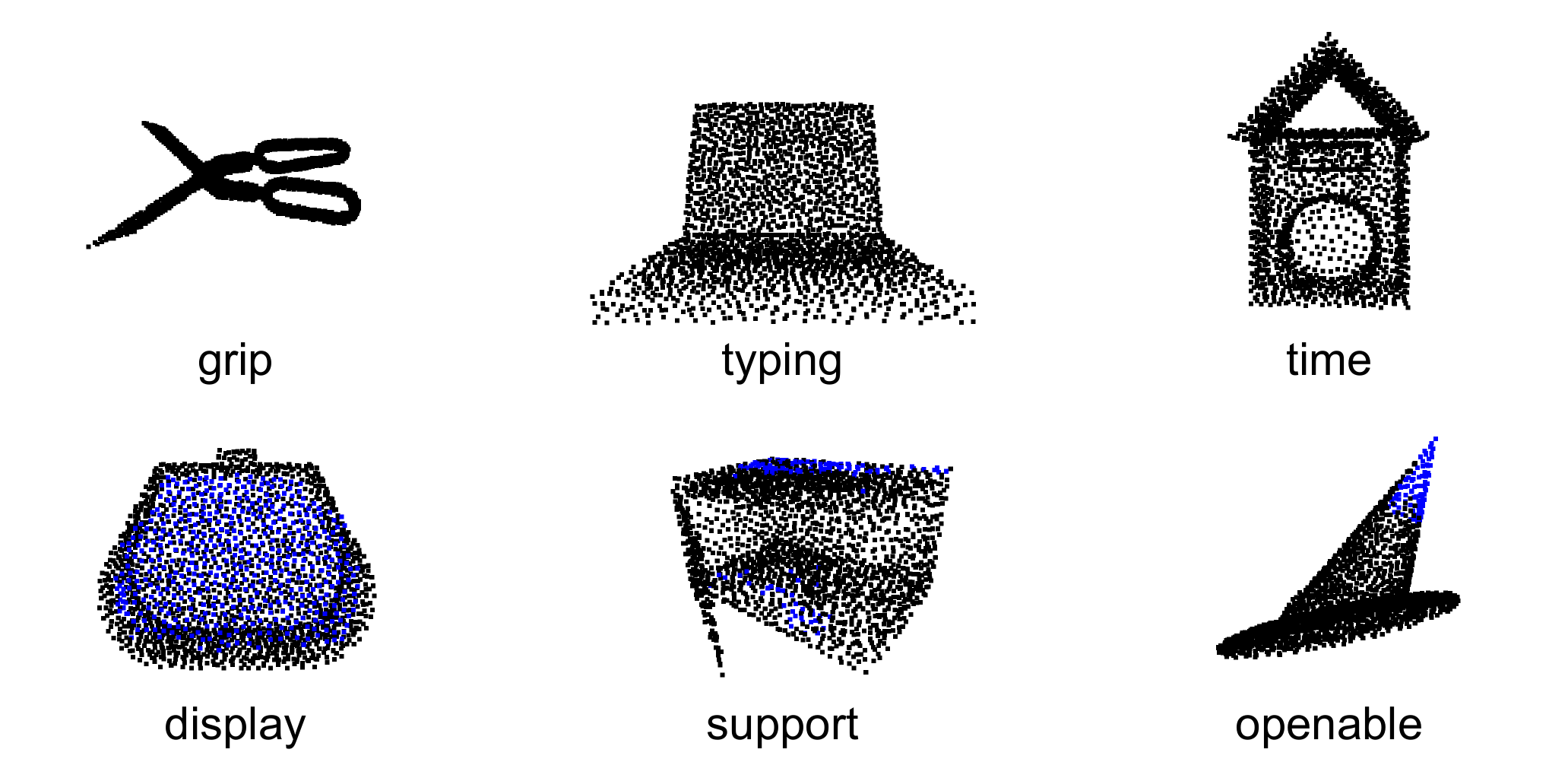}
 \vspace{1pt}
	\caption{Failure cases of OpenAD. \textit{Upper row:} Cases when OpenAD fails to detect unseen affordances. \textit{Lower row:} OpenAD detects affordances that are not furnished by the objects. }
	\label{fig:failure_cases}
\end{figure}

Despite achieving promising results, OpenAD has its limitation. Our method is still far from being able to detect completely unseen affordances.
The upper row of Figure~\ref{fig:failure_cases} shows cases when OpenAD fails to detect unseen affordances on seen objects.
Moreover, we present false-positive predictions of OpenAD in the lower row of Figure~\ref{fig:failure_cases}. In these cases, OpenAD detects affordances that the objects do not provide, i.e., \texttt{display} for the bag, \texttt{support} for the microwave, and \texttt{openable} for the hat.
From our intensive experiments, we see several improvement points for future work: \textit{i)} learning the visual-language correlation plays an important role in this task and can be further improved by using more complicated
techniques such that cross-attention mechanism~\cite{vaswani2017attention}
, \textit{ii)} applying a stronger point cloud backbone would likely improve the result, and \textit{iii)} having a large-scale dataset with several affordance classes would be beneficial for benchmarking and real-world robotic applications. Finally, we will also release our source code to encourage further study.

\section{Conclusions}\label{Sec:con}
We proposed OpenAD, a simple yet effective method for open-vocabulary affordance detection in 3D point clouds. Different from traditional approaches, OpenAD, with its capability of semantic understanding, can effectively detect unseen affordances without requiring annotated examples. Empirical results show that OpenAD outperforms other methods by a large margin. We further verified the capability of OpenAD to detect unseen affordances on both known and unseen objects. We additionally demonstrated the usability of OpenAD in real-world robotic applications.  
Although there is ongoing work to be accomplished, OpenAD's results provide encouraging evidence that intelligent robots understand many possibilities and perform better in complex environments.

\bibliographystyle{class/IEEEtran}
\bibliography{class/IEEEabrv,class/reference}
   
\end{document}